\documentclass[information,article,accept,oneauthor,pdftex,12pt,a4paper]{mdpi}

\usepackage{comment}
\usepackage{fancyvrb}
\usepackage{graphicx}
\usepackage{ragged2e}
\usepackage{hyperref}
\usepackage{soul}

\lastpage{x}
\doinum{10.3390/------}
\pubvolume{xx}
\pubyear{2013}
\history{Received: 26 May 2013; in revised form: 13 December 2013 / Accepted: 13 December 2013 / \\Published: xx}

\Title{The SP Theory of Intelligence: Benefits and Applications}

\Author{J Gerard Wolff}

\address[1]{CognitionResearch.org, Menai Bridge LL59 5LR,
UK; E-Mail:~jgw@cognitionresearch.org; \linebreak Tel.: +44-1248-712962; +44-7746-290775}

\corres{}

\abstract{This article describes existing and expected benefits of the \textit{SP theory of intelligence},
and some potential applications. The theory aims to simplify and integrate ideas across artificial intelligence, mainstream computing, and human perception and cognition, with information compression as a unifying theme. It combines conceptual simplicity with descriptive and explanatory power across several areas of computing and cognition. In the \textit{SP machine}---an expression of the SP theory which is currently realized in the form of a computer model---there is potential for an overall simplification of computing systems, including software. The SP theory promises deeper insights and better solutions in several areas of application including, most notably, unsupervised learning, natural language processing, autonomous robots, computer vision, intelligent databases, software engineering, information compression, medical diagnosis and big data. There is also potential in areas such as the semantic web, bioinformatics, structuring of documents, the detection of computer viruses, data fusion, new kinds of computer, and the development of scientific theories. The theory promises seamless integration of structures and functions within and between different areas of application. The potential value, worldwide, of these benefits and applications is at least \$190 billion each year. Further development would be facilitated by the creation of a high-parallel, open-source version of the SP machine, available to researchers everywhere.
}


\begin{document}

\newpage

\section{Introduction}


The {\em SP theory of intelligence} aims to simplify and integrate concepts across artificial intelligence, mainstream computing and human perception and cognition, with information compression as a unifying theme. This article describes existing and expected benefits of the SP theory and some of its potential~applications.

The theory is described most fully in \cite{wolff_2006} and more briefly in an extended overview \cite{sp_extended_overview}. This article should be read in conjunction with either or both of those accounts.

In brief, the existing and expected benefits of the theory are:

\begin{itemize}

\item Conceptual simplicity combined with descriptive and explanatory power.
\item Simplification of computing systems, including software.
\item Deeper insights and better solutions in several areas of application.
\item Seamless integration of structures and functions within and between different areas of application.

\end{itemize}

These points will be discussed, each in its own section, below. But first, as a background and context to what follows, there is a little philosophy, and an introduction to the SP theory with some associated~ideas.

\section{A Little Philosophy}\label{philosophy_section}

Amongst alternative criteria for the success of a theory, what appears to be the most widely accepted is the principle (``Occam's Razor'') that a good theory should combine {\em simplicity} with descriptive or explanatory {\em power} \cite{note_1}.
This equates with the idea---which echoes the underlying theme of the SP theory itself---that a good theory should compress empirical data via a reduction of ``redundancy'' in the data (thus increasing its ``simplicity''), whilst retaining as much as possible of its non-redundant descriptive ``power'' \cite{note_2}.
As John Barrow has written: ``Science is, at root, just the search for compression in the world.'' \cite{barrow_1992} (p. 247).

These principles are prominent in most areas of science: the Copernican heliocentric theory of the Sun and the planets is rightly seen to provide a simpler and more powerful account of the facts than Ptolemy's Earth-centered theory, with its complex epicycles; physicists have a keen interest in the possibility that quantum mechanics and relativity might be unified; biology would be greatly impoverished without modern understandings of evolution, genetics, and DNA; and so on. But in research in computer science, including artificial intelligence, there is a fragmentation of the field into a myriad of concepts and many specialisms, with little in the way of an overarching theory to pull everything together \cite{note_3}.

Perhaps the concept of a universal Turing machine \cite{turing_1936, note0} provides what is needed? That theory has of course been brilliantly successful---a basis for the current generation of computers and the many useful things that they can do. But it does not solve the problem of fragmentation and, although Alan Turing recognized that computers might become intelligent \cite{turing_1950}, the Turing theory, in itself, does not tell us~how!

Are we getting closer to that goal? Yes, certainly. But as Hubert Dreyfus has suggested (\cite{dreyfus_1972} (p. 12) and \cite{dreyfus_1992}), we may be getting closer to achieving real intelligence in computers in the same sense that a man climbing a tree is getting closer to reaching the moon.

Whether or not the SP theory succeeds in getting us to the AI moon, I believe there is a pressing need, in computer science and AI, for a stronger focus on the simplification and integration of ideas. In all areas of application, including quite specialized areas, a theory that scores well in terms of simplicity and power is, compared with any weaker theory, likely to yield deeper insights, better solutions, and better integration of structures and functions, both within a given area and amongst different areas (see also Sections \ref{areas_of_application_section} and \ref{integration_section}).

\section{The SP Theory and the SP Machine: A Summary}\label{sp_summary_section}

\begin{figure}[H]
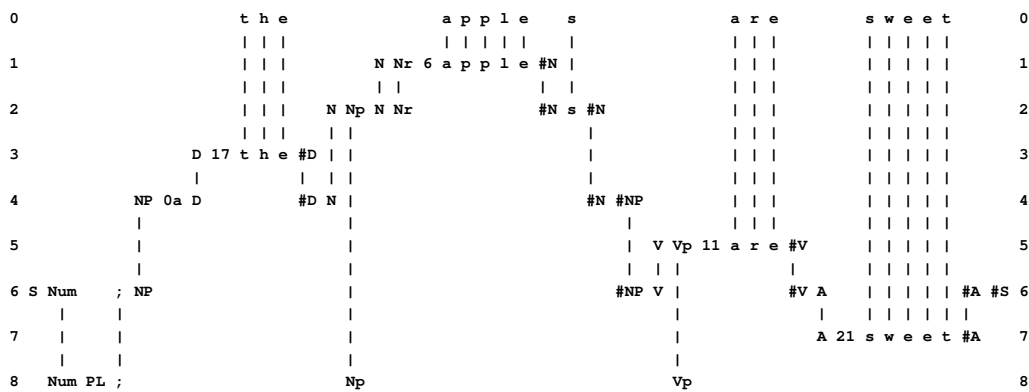
\vspace{6pt}
\fontsize{06.00pt}{07.20pt}
\centering
{\bf
\begin{BVerbatim}
0                       t h e                a p p l e    s                a r e         s w e e t       0
                        | | |                | | | | |    |                | | |         | | | | |
1                       | | |         N Nr 6 a p p l e #N |                | | |         | | | | |       1
                        | | |         | |              |  |                | | |         | | | | |
2                       | | |    N Np N Nr             #N s #N             | | |         | | | | |       2
                        | | |    | |                        |              | | |         | | | | |
3                  D 17 t h e #D | |                        |              | | |         | | | | |       3
                   |          |  | |                        |              | | |         | | | | |
4            NP 0a D          #D N |                        #N #NP         | | |         | | | | |       4
             |                     |                            |          | | |         | | | | |
5            |                     |                            |  V Vp 11 a r e #V      | | | | |       5
             |                     |                            |  | |           |       | | | | |
6 S Num    ; NP                    |                           #NP V |           #V A    | | | | | #A #S 6
     |     |                       |                                 |              |    | | | | | |
7    |     |                       |                                 |              A 21 s w e e t #A    7
     |     |                       |                                 |
8   Num PL ;                       Np                                Vp                                  8
\end{BVerbatim}
}
\caption{The best multiple alignment created by the SP computer model with a store of Old patterns like those in rows 1 to 8 (representing grammatical structures, including words) and a New pattern (representing a sentence to be parsed) shown in row 0.}
\label{parsing_1_figure}
\end{figure}

In broad terms, the SP theory has three main elements:

\begin{itemize}

\item All kinds of knowledge are represented with {\em patterns}: arrays of atomic symbols in one or two~dimensions.

\item At the heart of the system is compression of information via the matching and unification (merging) of patterns, and the building of {\em multiple alignments} like the one shown in Figure \ref{parsing_1_figure}.

\item The system learns by compressing ``New'' patterns to create ``Old'' patterns like those shown in rows~1 to 8 in Figure \ref{parsing_1_figure}.

\end{itemize}
\vspace {-6pt}

\begin{figure}[H]
 \centering
\includegraphics[width=0.8\textwidth]{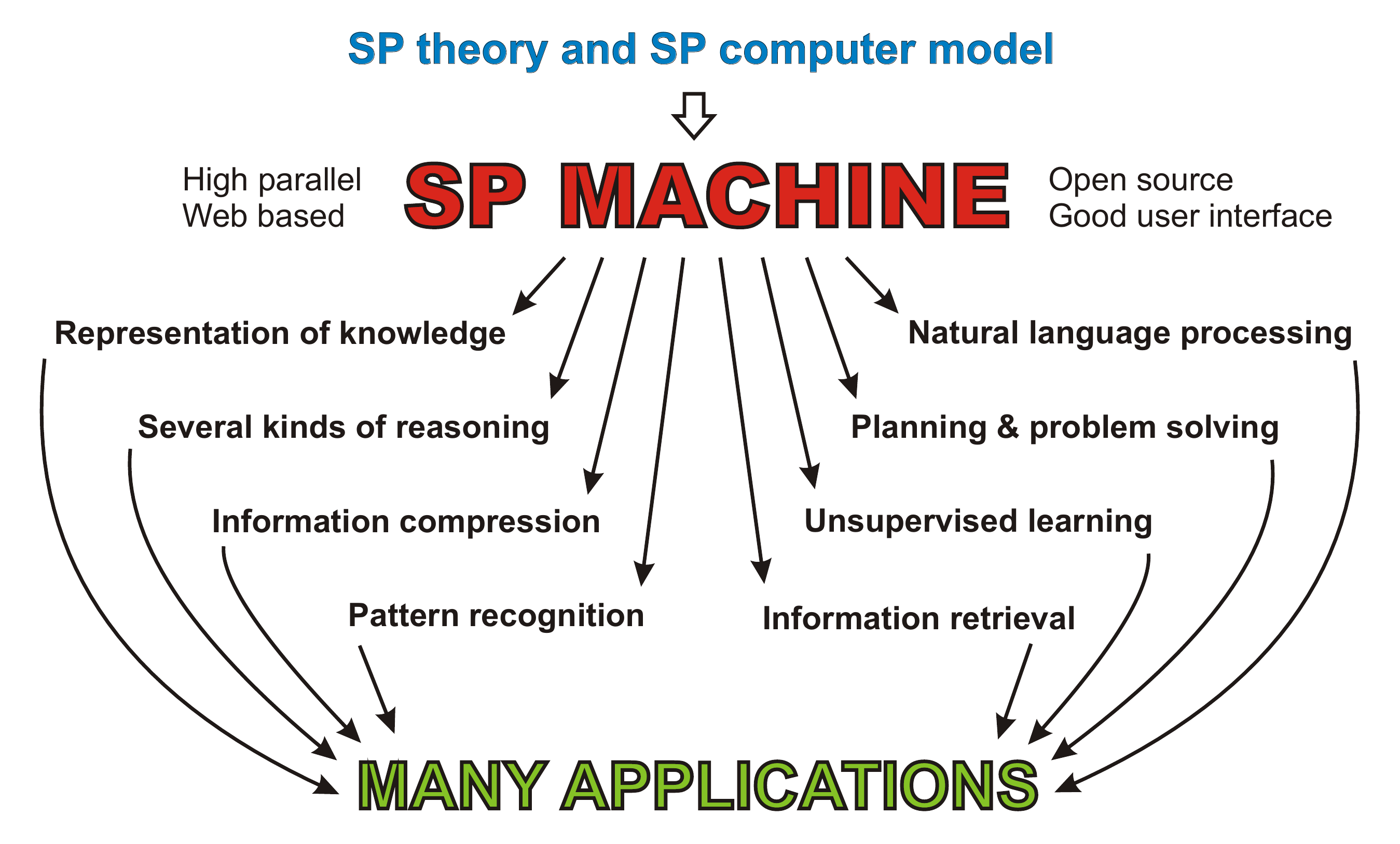}
\caption{Schematic representation of the development and application of the proposed SP machine. Reproduced with permission from Figure 2 in \cite{sp_extended_overview}.}
\label{sp_machine_figure}
\end{figure}

The system is realized in the form of a computer model \cite{note_4}.
It is envisaged that this will be developed into a high-parallel, open-source {\em SP machine}, an expression of the SP theory and a means for it to be applied (Section 3.2 in \cite{sp_extended_overview}). It is envisaged that this will be a facility for researchers everywhere to investigate what can be done with the system and create new versions of it. How things may develop is shown schematically in Figure \ref{sp_machine_figure}.
\pagebreak

\section{Combining Conceptual Simplicity with Descriptive and Explanatory Power}\label{sp_simplicity_power_section}

In principle, it should be possible to evaluate scientific theories quite precisely in terms of conceptual simplicity and descriptive and explanatory power---and to compare one theory with another---by measuring their ability to compress empirical data. But, for much of science, techniques and technologies are not yet well enough developed to make this feasible, and there is in any case the difficulty that it is rare for rival theories to address precisely the same body of empirical data. So in evaluating theories we normally need to rely on more-or-less informal judgments of simplicity and power (see also Section~6.10.7).

In those terms, the SP theory appears to score well:

\begin{itemize}

\item The SP theory, including the multiple alignment concept, is not trivially simple but it is not unduly complicated either. The SP computer model (Sections 3.9, 3.10 and 9.2 in \cite{wolff_2006}; Section 3.1 in \cite{sp_extended_overview}), which is, apart from some associated thinking, the most comprehensive expression of the theory as it stands now, is embodied in an ``exec'' file requiring less than 500 KB of storage space.

\item Largely because of the versatility of the multiple alignment concept, the SP theory can model a range of concepts and phenomena in: unsupervised learning, concepts of ``computing'', mathematics and logic, the representation of knowledge, natural language processing, pattern recognition, several kinds of probabilistic reasoning, information storage and retrieval, planning and problem solving, and information compression; and it has things to say about concepts in neuroscience and in human perception and cognition. (Chapters 4--13 in~\cite{wolff_2006}; Sections 5--13 in~\cite{sp_extended_overview}).

\end{itemize}

\newpage

\section{Simplification of Computing Systems, Including Software}\label{simplification_section}

Apart from the simplification and integration of concepts in artificial intelligence, mainstream computing, and human perception and cognition (Section \ref{sp_simplicity_power_section}), the SP theory can help to simplify computing systems, including software.

The principle to be described here is already familiar in the way databases and expert systems are~structured.

Early databases were each written as a monolithic system containing everything needed for its operation, and likewise for early expert systems. But it soon became apparent that, in the case of databases, a lot of effort could be saved by creating a generalized ``database management system'' (DBMS), with a user interface and mechanisms for the storage and retrieval of information, and then creating new databases by loading the DBMS with different bodies of information, according to need. In a similar way, an ``expert system shell'', with a user interface and with mechanisms for inference and for the storage and retrieval of information, eliminates the need to create those facilities repeatedly in each new expert system.

The SP system takes this principle further. It aims to provide a general-purpose \linebreak ``intelligence''---chiefly the multiple alignment framework---and thus save the need to create those kinds of mechanisms repeatedly in different AI applications: pattern recognition, natural language processing, several kinds of reasoning, planning, problem solving, unsupervised learning, and more.

The principle extends to conventional applications, since the matching and unification of patterns and a process of searching amongst alternative matches for patterns (which are elements of the multiple alignment framework), are used in many kinds of application outside the world of artificial intelligence.

\begin{figure}[!h]
\centering
\includegraphics[width=0.8\textwidth]{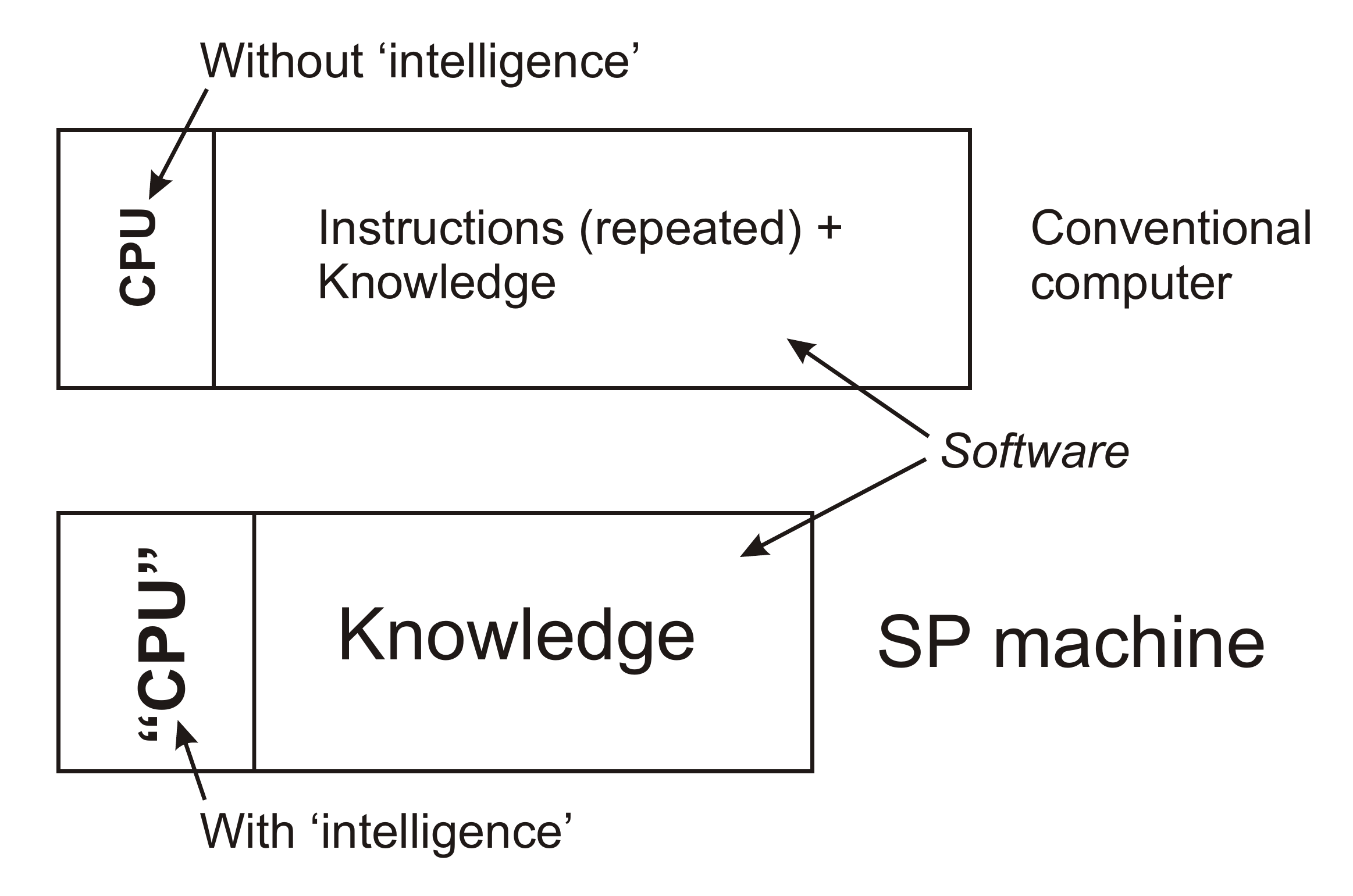}
\caption{Schematic representations of a conventional computer and an SP machine, as discussed in the text. Adapted with permission from Figure 4.7 in \cite{wolff_2006}.}
\label{computers_schematic_figure}
\end{figure}

The way in which this principle may help to simplify computing systems is shown schematically in Figure \ref{computers_schematic_figure}. In a conventional computer, shown at the top of the figure, there is a central processing unit~(CPU) which is relatively simple and without ``intelligence''. Software in the system, shown to the right, is a combination of two things:

\begin{itemize}

\item Domain-specific knowledge such as knowledge of accountancy, geography, the organization and procedures of a business, and so on \cite{note_6}.

\item Processing instructions to provide the intelligence that is missing in the CPU, chiefly processes for the matching and unification of patterns, and for searching amongst alternative matches to find one or more that are ``good''. These kinds of processing instructions---let us call them MUP instructions---are used in many different kinds of application, meaning that there is a considerable amount of redundancy in conventional computing systems---where the term ``system'' includes both the variety of programs that may be run, as well as the hardware on which they run.

\end{itemize}

In the SP machine, shown schematically at the bottom of the figure, the ``CPU'' \cite{note_6a} is more complicated (as suggested by the larger size of ``CPU'' in the figure) and aims to provide much of the ``intelligence'' that is missing in the CPU of a conventional computer. This should mean that MUP instructions, and their many repetitions, may be largely eliminated from the software, leaving just the domain-specific knowledge. The increased complexity of the ``CPU'' should be more than offset by cutting out much of the redundancy in the software, meaning an overall reduction in complexity---as suggested by the relatively small size of the figure that represents the SP machine, compared with the figure that represents a conventional computer.

An overall simplification of computing systems can bring benefits like these:

\begin{itemize}

\item {\em Savings in development effort and associated costs}. With more intelligence in the CPU there should be less need for it to be encoded in applications.

\item {\em Savings in development time}. With a reduced need for hand crafting, applications may be developed more quickly.

\item {\em Savings in storage costs}. There may be useful economies in the storage space required for application code.

\item Related benefits noted in Section \ref{information_compression_section}.

\end{itemize}

\section{Deeper Insights and Better Solutions in Several Areas of Application}\label{areas_of_application_section}

This section describes some potential areas of application for the SP theory, and its potential benefits, drawing on the capabilities of the SP system (described in \cite{wolff_2006} (Chapters 4--11) and \cite{sp_extended_overview} (Sections 5--13)). These examples are probably just the tip of the iceberg. To mix our metaphors, there is a rich vein to be~mined.

Because of its importance, I believe it is worth repeating the principle, mentioned in Section \ref{philosophy_section}, that, even in specialized areas, a theory like the SP theory that has a broad base of support is likely to provide deeper insights than others, better solutions to problems, and (Section \ref{integration_section}) better integration of structures and functions within and between different areas of application. Putative examples from the SP theory~include:

\begin{itemize}

\item A new method for encoding discontinuous dependencies in syntax (Chapter~5 in \cite{wolff_2006}; Section 8.1 in \cite{sp_extended_overview}), which is arguably simpler and more versatile than older methods.

\item The multiple alignment framework, with the use of one simple format for all kinds of knowledge, is likely to facilitate the seamless integration of the syntax and semantics of natural languages, with the kinds of subtle interaction described in Section \ref{nl_applications_section}, below.

\item The multiple alignment framework provides for the seamless integration of class inclusion hierarchies and part-whole hierarchies (Section 6.4 in \cite{wolff_2006}; Section 9.1 in \cite{sp_extended_overview}), something which is missing in most object-oriented systems.

\item The use of one framework for both the analysis and production of knowledge (Section 3.8 in~\cite{wolff_2006}; Section 4.5 in \cite{sp_extended_overview}) is a significant simplification compared with systems in which there are separate~mechanisms.

\item The multiple alignment framework provides for the integration of the relational, object-oriented, and network models for databases, with improvements in the relational model \cite{wolff_sp_intelligent_database}.

\item The multiple alignment framework, with the use of one simple format for all kinds of knowledge, is likely to facilitate the learning of concepts that include diverse kinds of knowledge.

\end{itemize}

Theories with a narrow focus may work reasonably well in the target area but are likely to fail when the scope is widened. Even if existing technologies in a given area are doing well, there is likely to be a case for examining how the SP theory may be applied, and its potential benefits.

These kinds of benefits of a good theory have been seen repeatedly in the history of science: the germ theory of disease serves us better than the idea that diseases are caused by ``bad air''; recognizing the role of oxygen in combustion is more practical than the phlogiston theory; space travel does not work unless we understand the laws of physics; and so on.

\subsection{Unsupervised Learning}

The first area of application to be considered---unsupervised learning---is a bit different from the others because it is less an area of application in its own right and more a contribution to others, most notably autonomous robots (Section \ref{autonomous_robots_section}), computer vision (Section \ref{computer_vision_section}), software engineering (Section~\ref{software_engineering_section}), big data (Section \ref{big_data_section}), the semantic web (Section 6.10.1),
bioinformatics (Section 6.10.2),
and the development of scientific theories (Section 6.10.7).

Research developing computer models of language learning
(reviewed in \cite{wolff_1988}; see also \cite{note_6b})---in particular, the importance of information compression in that area---has been one of the main sources of inspiration for the SP theory. But radical restructuring has been needed to meet the goal of simplifying and integrating ideas across artificial intelligence, mainstream computing, and human perception and~cognition.

The new model, SP70 \cite{sp_extended_overview} (Section 3.1), has already demonstrated an ability to discover generative grammars from samples of artificial languages, including segmental structure, classes of structure, and abstract patterns. But some further work is needed to realize its full potential for learning, as outlined in~\cite{sp_extended_overview} (Section 3.3). I believe the residual problems are soluble, and that, with their solution, the SP machine will have considerable potential in the areas mentioned above, and others.

A particular area of interest is, of course, the learning of natural languages, perhaps starting with the learning of syntactic structures. On longer time scales, their is potential for the learning of non-syntactic ``semantic'' structures and their integration with syntax.

\subsection{Applications in the Processing of Natural Language}\label{nl_applications_section}

As noted in \cite{sp_extended_overview} (Section 8), the SP system supports the parsing and production of language, it provides a simple but effective means of representing discontinuous dependencies in syntax, it is robust in the face of errors in data, it facilitates the integration of syntax with semantics, and it provides a role for context in the processing of language. There are many potential applications, including such things as checking for grammatical errors in text; predictive typing that takes account of grammar as well as the spellings of individual words; natural-language front-ends to databases and expert systems; and the understanding and translation of natural languages.

It is true that some of these things are already done quite well with existing technologies but, as suggested in Sections \ref{philosophy_section}, \ref{areas_of_application_section}, and \ref{integration_section}, there may nevertheless be advantages in exploring what can be done with SP system. For example, there are some subtle interactions between syntax and semantics which which may be difficult to handle without seamless integration of the two:

\begin{itemize}

\item Compare ``Fafnir, who plays third base on the Little League team, is a fine boy.'' with ``They called their son Fafnir, which is a ridiculous name.'' \cite{mccawley_1968} (p.~139). The choice of ``who'' or ``which'' in these two sentences---a syntactic issue---depends on the {\em meaning} of ``Fafnir''. In the first sentence, it means a person, while in the second sentence it means a name.

\item As a contrasting example, compare ``John gave me the scissors; I am using them now.'' with ``John gave me the two-bladed cutting instrument; I am using it now.'' \cite{mccawley_1968} (p.~141). In this case, the choice of ``them'' or ``it''---also a syntactic issue---depends on the {\em syntactic form} of what is referred to: ``the scissors'' has a plural form while ``the two-bladed cutting instrument'' has a singular form, although the two expressions have the same meaning.

\end{itemize}

Apart from the integration of syntax with semantics, the SP system is likely to smooth the path for the integration of natural language processing with other aspects of intelligence: reasoning, pattern recognition, problem solving, and so on.

\subsubsection{Towards the Understanding and Translation of Natural Languages}\label{crowdsourcing_section}

Given the apparent simplicity of representing syntactic and non-syntactic knowledge with patterns~\cite{sp_extended_overview} (Sections 7 and 8), it may be feasible to use crowd-sourcing in the manner of Wikipedia to create syntactic rules for at least one natural language and a realistically large set of ontologies to provide associated meanings. Existing grammars and existing ontologies may also prove useful in the creation of appropriate sets of SP patterns.

With this kind of knowledge on the web and with a high-parallel, on-line version of the SP machine (as sketched in \cite{sp_extended_overview} (Section 3.2)), it should be possible to achieve ``understanding'' of natural language, either in restricted domains (Sections 6.2.2, 6.2.3, and 6.2.4, below)
or more generally.

With the same ontologies and with syntactic knowledge for two or more languages, it may be feasible to achieve translations between languages, with the ontologies functioning as an ``interlingua'' \cite{note_7}.

The use of an interlingua should yield two main benefits:

\begin{itemize}

\item When translations are required amongst several different languages, it is only necessary to create $n$ mappings, one between each of the $n$ languages and the interlingua. This contrasts with \linebreak syntax-to-syntax systems, where, in the most comprehensive application of this approach, $n! / 2(n - 2)!$ pairs of languages are needed \cite{note_8}.

\item It seems likely that the use of an interlingua will be essential if machine translation is ever to reach or exceed the standard that can be achieved by a good human translator.

\end{itemize}

\subsubsection{Natural Language and Information Retrieval}\label{nl_information_retrieval_section}

The task of developing a system for understanding natural language may be considerably simplified if we narrow our focus to specific areas: those that require only a relatively restricted range of language forms. Three possibilities are outlined in this and the following two subsections.

On relatively short timescales the SP system may help to overcome some weaknesses in Internet search engines as they are now. For example, if we enter the query ``telephone extensions in Anglesey'' into any of the leading search engines, we get a mixture of answers: some of them---which are relevant to the query---relate to telephone extensions, while others---which are not relevant---relate to home extensions, building work and the like.

With the SP system, it should be relatively straightforward to overcome that kind of problem, at least for queries about businesses or services in a given area:

\begin{itemize}

\item Since a usefully-large range of natural-language questions may be asked with fairly simple language, it would only be necessary---for any given language---to create a relatively simple set of syntactic rules and lexicon, with corresponding semantic structures, all of them expressed with SP~patterns.

\item It should be possible to create an online tool that would make it easy for the owner or manager of a business to create an ontology for that business that describes what it is, what it does, and the kinds of products or services that it sells. It would of course be necessary for many businesses to provide that kind of information.

\item With those two things in place, it should be possible, via the SP system, to create a semantic analysis for a given query (as outlined in \cite{wolff_2006} (Section 5.7)) and to use that analysis to identify businesses or services that satisfy the query.

\end{itemize}

With this approach, a query like ``telephone extensions in Anglesey'' would be parsed into constituents like ``telephone extensions'' and ``in Anglesey'', each with a semantic structure. The word ``extensions'', with a buildings-related meaning, would not figure in those semantic structures, so the unhelpful responses to the query would be eliminated.

\subsubsection{Going Beyond FAQs}\label{beyond_FAQs_section}

Many websites provide a list of ``frequently asked questions'' or FAQs which, as their name suggests, attempt to provide answers to the more commonly-asked questions about the subject of the website. This is OK as far as it goes but it would clearly be more useful if users could simply ask questions in the normal way and receive sensible answers.

To do this with the SP system means developing an ontology for the subject of the website, with an associated vocabulary and syntax, all expressed with SP patterns. This should be feasible to do, partly because of the essential simplicity of representing knowledge in the SP system, and partly because it is likely that the subject matter will be circumscribed.

\subsubsection{Interactive Services}\label{interactive_services_section}

In a similar way, it should be possible to develop appropriate ontologies, vocabularies and syntaxes for such applications as booking a hotel room, booking a place on a plane or a train, booking a seat in a restaurant, and so on.

\subsection{Towards a Versatile Intelligence for Autonomous Robots}\label{autonomous_robots_section}

If a robot is to survive and be effective in a challenging environment like Mars, where communication lags restrict the help that can be provided by people, there are advantages if it can be provided with human-like intelligence and versatility---as much as possible. A theory like the SP theory, which provides a unified view of several different aspects of intelligence, is a good candidate for the development of that general intelligence \cite{note_9}.

The SP system has the potential to give robots human-like flexibility and adaptability. Consider, for example, how top-ranking players learn to play games like pool, billiards, or snooker. This is quite different from how some existing robots play pool (see, for example, \cite{archibald_etal_2010}), where the situation is simplified compared with what a human player faces, and where the robot is provided with a knowledge of geometry and associated mathematics.

For a person to play that kind of game well, they must spend thousands of hours on practice. Potting a ball means visual perception---from the position of the player, not an overhead camera---of the table, the cue ball, the target ball, the pocket and the cue, together with a knowledge of the motor patterns involved in striking the cue ball, and how those motor patterns are shaped by feedback from muscles and from sensory information about the impact between the cue and the cue ball.

Amongst the thousands of such configurations that a player may encounter, he or she must learn which ones lead to success. Getting beyond mere potting of a ball means learning how to impart spin to the cue ball, and judging such things as timings and forces, so that, after a shot, the configuration of balls on the table is favorable for the next shot by the given player, or  difficult for his or her opponent. As before, there are thousands of associations to be learned, in this case between visual and motor patterns when the cue ball is struck and the configuration of the balls on the table when they have all stopped moving.

At first sight, the pool-playing robot seems to have the edge over the human player because it does not need all those thousands of hours of practice. But what is missing is that afore-mentioned flexibility and adaptability: without task-specific programming, the robot cannot do such things as make a cup of tea or take the dog for a walk. By contrast, most people can learn to do those kinds of things without~difficulty.

In this connection, the main strengths of the SP system are:

\begin{itemize}

\item Potential for the kind of visual analysis needed to assimilate the many configurations of balls, pockets, and cue (Section \ref{computer_vision_section}, next).

\item The versatility of the SP framework in the representation and processing of diverse kinds of knowledge (Section \ref{integration_section}) should facilitate the seamless integration of visual information about the table, balls, and so on, with information about actions by the player and feedback from muscles and from touch.

\item Via its processes for unsupervised learning \cite{sp_extended_overview} (Section 5), the SP system should be able to learn associations between visual, motor and feedback patterns, as described above. It appears that similar principles have potential in the kinds of everyday learning mentioned above.

\end{itemize}

\subsection{Computer Vision}\label{computer_vision_section}

The SP system has potential in the area of computer vision, as described in \cite{sp_vision}. In brief:

\begin{itemize}

\item It has potential to simplify and integrate several areas in computer vision, including feature detection and alignment, segmentation, deriving structure from motion, stitching of images together, stereo correspondence, scene analysis, and object recognition (see, for example, \cite{szeliski_2011}). With regard to the last two topics:

\begin{itemize}

\item Since scene analysis is, in some respects, similar to the parsing of natural language, and since the SP system performs well in parsing, it has potential in scene analysis as well. In the same way that a sentence may be parsed successfully without the need for explicit markers of the boundaries between successive words or between successive phrases or clauses, a scene may be analyzed into its component parts without the need for explicit boundaries between objects or other elements in a scene.

\item The system provides a powerful and flexible framework for pattern recognition, as outlined in \cite{sp_extended_overview} (Section 9), and more fully described in \cite{wolff_2006} (Chapter 6). It seems likely that this can be generalized for object recognition.

\end{itemize}

\item The SP system has potential for unsupervised learning of the knowledge required for recognition. For example, discrete objects may be identified by the matching and merging of patterns within stereo images ({\em cf.} \cite{note_10, marr_poggio_1979})
    or within successive frames in a video, in much the same way that the word structure of natural language may be discovered via the matching and unification of patterns~\cite{sp_extended_overview} (Section 5.2). The system may also learn such things as classes of entity, and associations between entities, such as the association between black clouds and rain.

\item The system is likely to facilitate the seamless integration of vision with other aspects of intelligence: reasoning, planning, problem solving, natural language processing, and so on.

\item As noted in \cite{sp_extended_overview} (Sections 8 and 9), the system is robust in the face of errors of omission, commission or substitution---an essential feature of any system that is to achieve human-like capabilities in vision.

\item With regard to those problems outlined in \cite{sp_extended_overview} (Section 3.3), that relate to vision, there are potential~solutions:

\begin{itemize}

\item It is likely that the framework can be generalized to accommodate patterns in two dimensions.

\item As noted in \cite{sp_extended_overview} (Section 3.3), 3D structures may be modeled using 2D patterns, somewhat in the manner of architects' drawings. Knowledge of such structures may be built via the matching and unification of partially-overlapping 2D views \cite{sp_vision} (Sections 6.1 and 6.2). Although such knowledge, as with people, may be partial and not geometrically accurate~\cite{note_11},
    it can, nevertheless, be quite serviceable for such purposes as getting around in one's environment.

\item The framework has potential to support both the discovery and recognition of low-level perceptual features.

\end{itemize}

\end{itemize}

\subsection{A Versatile Model for Intelligent Databases}\label{intelligent_databases_section}

As described in \cite{wolff_sp_intelligent_database} and outlined in \cite{sp_extended_overview} (Section 11), the SP system provides a versatile framework that accommodates several models of data (including object-oriented, relational, network and tree models), that facilitates the integration of different models, and that provides capabilities in reasoning, learning, pattern recognition, problem-solving, and other aspects of intelligence.

An industrial-strength version of the SP machine may prove useful in several areas, including, for example, crime investigations:

\begin{itemize}

\item The system would provide a means of storing and managing the data that are gathered in such investigations, often in large amounts.

\item It may help in the recognition of features or combinations of features that link a given crime to other crimes, either current or past---and likewise for suspects.

\item The system's capabilities in pattern recognition may also serve in the scanning of data to recognize indicators of criminal activity.

\item It may prove useful in piecing together coherent patterns from partially-overlapping fragments of information, in much the same way that partially-overlapping digital photographs may be stitched together to create a larger picture.

\item Given the capabilities of the SP system for different kinds of reasoning (Chapter 7 in \cite{wolff_2006} and Section 10 in \cite{sp_extended_overview}), including reasoning with information that is not complete, the system may prove useful in suggesting avenues to be explored, perhaps highlighting aspects of an investigation that might otherwise be overlooked.

\item Transparency in the representation of knowledge \cite{sp_extended_overview} (Section 2.4) and in reasoning via the multiple alignment framework (Chapters 7 and 10 in \cite{wolff_2006} and Section 10 in \cite{sp_extended_overview}) are likely to prove useful where evidence needs to be examined in a court of law, an advantage over ``sub-symbolic'' systems that yield results without insight.

\end{itemize}

\subsection{Software Engineering}\label{software_engineering_section}

The SP system has potential in software engineering, as described in the following subsections.

\subsubsection{Procedural Programming}\label{procedural_programming_section}

Procedural programming of a traditional kind is not merely an artifact of the way computers have been designed. It provides an effective means of modeling the kinds of procedures in the real world where the sequence of actions is important. For example, if we are making a cake, we need to mix the ingredients before putting the cake in the oven, and we need to bake the cake before icing it.

It appears that the SP system provides much of what is needed to model real-world procedures of that~kind:

\begin{itemize}

\item {\em Procedure}. A sequence of actions may be modeled with a \newline one-dimensional SP pattern such as ``\texttt{do\_A do\_B do\_C~...}''.

\item {\em Variables, values and types}. In the SP framework, any pair of neighboring symbols may function as a slot or variable into which other information may be inserted via the alignment of patterns. For example, the pair of symbols ``\texttt{Num ;}'' in row 6 of Figure \ref{parsing_1_figure} is, in effect, a variable containing the value ``\texttt{PL}'' from row 8. The ``type'' of the variable is defined by the range of values it may take, which itself derives from the set of Old patterns provided for the SP model when the multiple alignment was created. In this case, the variable ``\texttt{Num ;}'' may take the values ``\texttt{PL}'' (meaning ``plural'') or ``\texttt{SNG}'' (meaning ``singular'').

\item {\em Function or subroutine}. In the SP framework, the effect of a function or subroutine may be achieved in much the same way as with variables and values. For example, the pattern \linebreak ``\texttt{A 21 s w e e t \#A}'' in row 7 of Figure \ref{parsing_1_figure} may be seen as a function or subroutine that has been ``called'' from the higher-level procedure ``\texttt{S Num ; NP \#NP V \#V A \#A \#S}'' in row 6 of the figure.

\item {\em Function with parameters}. As can be seen in Figure \ref{parsing_2_figure}, the SP system may be run ``backwards'', much as can be done with an appropriately-designed Prolog program, generating ``data'' from ``results'' (Section 3.8 in \cite{wolff_2006} and Section 4.5 in \cite{sp_extended_overview}). In this example, the pattern \linebreak ``\texttt{S Num ; NP \#NP V \#V A \#A \#S}'' may be seen as a ``function'', and symbols like \linebreak ``\texttt{PL 0a 17 ...}'' in row 0 may be seen as ``values'' for the ``parameters'' of the function.

\item {\em Conditional statements}. The effect of the values just described is to select elements of the multiple alignment: ``\texttt{PL}'' selects the pattern ``\texttt{Num PL ; Np Vp}'', ``\texttt{0a}'' selects ``\texttt{NP 0a D \#D N \#N \#NP}'', and so on. Each of these selections achieves the effect of a conditional statement in a conventional program: ``If `\texttt{PL}' then choose `\texttt{Num PL ; Np Vp}' ''; ``If `\texttt{0a}' then choose \linebreak `\texttt{NP 0a D \#D N \#N \#NP}' ''; and so on.

\begin{figure}[H]
\fontsize{06.00pt}{07.20pt}
\centering
{\bf
\begin{BVerbatim}
0 S     PL      0a   17                    6                            11            21              #S 0
  |     |       |    |                     |                            |             |               |
1 S Num |  ; NP |    |                     |                   #NP V    |        #V A |            #A #S 1
     |  |  | |  |    |                     |                    |  |    |        |  | |            |
2    |  |  | |  |    |                     |                    |  V Vp 11 a r e #V | |            |     2
     |  |  | |  |    |                     |                    |    |              | |            |
3    |  |  | NP 0a D |        #D N         |                #N #NP   |              | |            |     3
     |  |  |       | |        |  |         |                |        |              | |            |
4    |  |  |       D 17 t h e #D |         |                |        |              | |            |     4
     |  |  |                     |         |                |        |              | |            |
5    |  |  |                     |    N Nr 6 a p p l e #N   |        |              | |            |     5
     |  |  |                     |    | |              |    |        |              | |            |
6    |  |  |                     N Np N Nr             #N s #N       |              | |            |     6
     |  |  |                       |                                 |              | |            |
7    |  |  |                       |                                 |              A 21 s w e e t #A    7
     |  |  |                       |                                 |
8   Num PL ;                       Np                                Vp                                  8
\end{BVerbatim}
}
\caption{The best multiple alignment created by the SP model with the same Old patterns as were used in the creation of the multiple alignment shown in Figure \ref{parsing_1_figure}, and a New pattern representing an encoding of a sentence, shown in row 0.}
\label{parsing_2_figure}
\end{figure}

\item {\em Repetition of procedures}. Conventional programs provide for the repetition of procedures with recursive functions and with statements like {\em repeat~...~until} and {\em do~...~while}. In the SP system, repetition may be encoded with recursive (self-referential) patterns like ``\texttt{X 1 a b c X 1 \#X \#X}'' which may be applied as shown in Figure \ref{recursive_multiple_alignment_figure}.

\begin{figure}[!htbp]
\fontsize{10.00pt}{12.00pt}
\centering
{\bf
\begin{BVerbatim}
0     a b c     a b c     a b c     a b c                    0
      | | |     | | |     | | |     | | |
1 X 1 a b c X 1 | | |     | | |     | | |              #X #X 1
            | | | | |     | | |     | | |              |
2           X 1 a b c X 1 | | |     | | |           #X #X    2
                      | | | | |     | | |           |
3                     X 1 a b c X 1 | | |        #X #X       3
                                | | | | |        |
4                               X 1 a b c X 1 #X #X          4
\end{BVerbatim}
}
\caption{The best multiple alignment produced by the SP model with the New pattern ``\texttt{a b c a b c a b c a b c}'' and multiple appearances of the Old pattern, ``\texttt{X 1 a b c X 1 \#X \#X}''.}
\label{recursive_multiple_alignment_figure}
\end{figure}

\item {\em Integration of ``programs'' and ``data''}. Databases provide the mainstay of many software engineering projects and, as described in Chapter 6 in \cite{wolff_2006}, Section 11 in \cite{sp_extended_overview} and \cite{wolff_sp_intelligent_database}, the SP system promises benefits in that area. In view of the versatility of SP patterns in the multiple alignment framework to represent varied kinds of knowledge \cite{sp_extended_overview} (Section 7), there is potential in software engineering projects for the seamless integration of ``programs'' with ``data''.

\item {\em Object-oriented design}. The SP system provides for object-oriented design with class \linebreak hierarchies (including cross-classification), part-whole hierarchies, and inheritance of attributes (Section~6.4 in \cite{wolff_2006}; Section~9.1 in \cite{sp_extended_overview}). In view of the importance of object-oriented design in software engineering, this capability in the SP system is likely to prove useful for applications in that area.

\end{itemize}

To help smooth the path to working in the SP framework, there may be a case, at least in the early stages, for providing ``syntactic sugar'' to make things look more like conventional programming.

\subsubsection{No Compiling or Interpretation}

A potential advantage of the SP system compared with conventional systems is that there would not be any need for compiling or interpretation, meaning the translation of source code into machine code. Any SP program would simply be a set of SP patterns that would be processed directly by the SP machine.

It is true that the current SP computer model is compiled in the conventional way, and the same is likely to be true of any versions of the SP machine that are hosted on a conventional computing system. But the processes of building multiple alignments and creating new knowledge structures does not require compiling or interpretation as normally understood.

\subsubsection{Sequential and Parallel Processing}

In considering the application of sequential and parallel processing in the SP system, we need to distinguish between the workings of the SP machine and real-world processes (such as baking a cake or managing an industrial plant) to which the SP system may be applied:

\begin{itemize}

\item {\em The workings of the SP machine}. In the projected SP machine \cite{sp_extended_overview} (Section 3.2), it is envisaged that, while the ordering of symbols would be respected when one pattern is matched with another, the process of matching two patterns may be done left-to-right, right-to-left, or with many symbols matched in parallel. Likewise, many pairs of patterns may be matched in parallel.

\item {\em Sequencing and parallelism in real-world processes}. Although the SP machine may be flexible in the way parallelism is applied to the matching of patterns, the system needs to respect sequencing and parallelism in the real-world processes to which the system may be applied. In broad terms, this may be achieved via the SP patterns themselves:

\begin{itemize}

\item A sequence of operations may be modeled via the left-to-right sequence of symbols in a one-dimensional SP pattern, with ``subroutines'' if required, as outlined in Section 6.6.1.

\item Since SP patterns at the ``top'' level are independent of each other, they may serve to model processes that may run in parallel.

\end{itemize}

    Naturally, a realistic model of any real-world project is likely to need additional information about such things as timings, dependencies between processes, and resources required.

\end{itemize}

\subsubsection{Automatic Programming}

With some improvements to the learning processes in the SP model (Section 9.5 in \cite{wolff_2006}; Section 5.1.4 in \cite{sp_extended_overview}), there is potential to by-pass at least some of the labor-intensive and costly hand-crafting of computer programs which is currently the mainstay of software engineering:

\begin{itemize}

\item {\em Learning of ontologies}. Knowledge of significant categories---people, buildings, vehicles, and so on---may be built up incrementally from information that users supply to the system. There is potential for the automatic structuring of such information in terms of class hierarchies and part-whole hierarchies.

\item {\em Learning of procedures}. In a similar way, procedural knowledge may be built up from information supplied by users, and, via learning, such procedural knowledge may be structured automatically and generalized \cite{note_12}.
    Likewise, for ``programming by demonstration'' in robotics \cite{note_13}.

\end{itemize}

Since knowledge in the SP system is intended to be transparent and comprehensible by people~\cite{sp_extended_overview} (Section 2.4), it should be relatively straightforward to make adjustments and corrections to learned~structures.

\subsubsection{Verification and Validation}

As was suggested in Section \ref{simplification_section}, the SP system may reduce or eliminate MUP instructions and their repetition in different applications. To the extent that that proves possible, we may reduce the opportunities for errors to be introduced. There is a corresponding potential to reduce or eliminate the need for ``verification'' in software development, and for improvements in the quality of software.

The SP machine also has potential in ``validation'': helping to ensure that what is being developed is what is required. If domain-specific knowledge, including the requirements for a software system, can be loaded directly into an SP machine, without the need for the traditional kind of programming, this will help to ensure that what the system does is what the users want. Users may have tighter control than is traditional over what a system does, with more freedom to make changes when required \cite{note_14}.

\subsubsection{Technical Debt}

By reducing the complexity of the software development process and by enabling users to have more direct control over how things work, the SP machine may also help to reduce or eliminate the problem of ``technical debt'', the way in which conventional systems can accumulate a backlog of ``housekeeping'' tasks and become unmanageably complex with the passage of time.

\subsection{Information Compression}\label{information_compression_section}

Given the central importance of information compression in the SP theory, one would naturally expect the SP machine to prove useful in that area of application. In effect, the multiple alignment framework serves, {\em inter alia}, to integrate three techniques for information compression: {\em chunking-with-codes}, {\em schema-plus-correction}, and {\em run-length coding} \cite{wolff_2006} (Section 2.2.9).

There is potential for the SP system to achieve relatively high levels of lossless compression for two main reasons:

\begin{itemize}

\item It is intended that, normally, the SP machine will perform a relatively thorough search of the space of alternative unifications of patterns and achieve correspondingly high levels of compression.

\item If, as anticipated with some further development, the system will be able to learn discontinuous dependencies in data (see \cite{sp_extended_overview} (Section 3.3)), it will tap into sources of redundancy that appear to be outside the scope of traditional methods for compression of information.

\end{itemize}

In general, information compression can bring benefits in: economies in storage; speeding up transmission of information or reducing the demand for bandwidth, or some combination of the two (Section 6.7.1); providing a basis for inference (as described in \cite{sp_extended_overview} (Sections 2.1, 4.4, 10), and elsewhere in that article); and greater efficiency in processing with corresponding gains in energy efficiency (Section 6.7.2).

\subsubsection{Economies in the Transmission of Information}\label{economical_transmission_section}

A compressed version of a body of information, $I$, may be seen to comprise two parts:

\begin{itemize}

\item A ``grammar'' for $I$ (which we may call $G$) containing patterns, at one or more levels of abstraction, that occur repeatedly in $I$.

\item An ``encoding'' of $I$ in terms of the grammar (which we may call $E$), including non-repeating information in $I$.

\end{itemize}

Where there is significant redundancy in $I$---which is true of most kinds of natural-language text and most kinds of images, especially moving images---$E$ is likely to be much smaller than $G$. A grammar for $I$ may also support the lossless compression of other bodies of information, provided they contain the same kinds of structures as $I$.

These things can provide the means of transmitting information very economically:

\begin{itemize}

\item A receiver (such as a TV set or a computer receiving web pages over the Internet) may be equipped with a grammar for the kind of information it is designed to receive, and some version of the SP~system.

\item Instead of transmitting ``raw'' data, or data that has been compressed in the traditional manner (containing both $G$ and $E$), the encoding by itself would be sufficient.

\item Using the SP system, the original data may be reconstructed fully, without any loss of information, by decoding the transmitted information ($E$) in terms of the stored copy of $G$.

\end{itemize}

A simple version of this idea, using a dictionary of words as a ``grammar'' for ordinary text, is already recognized (e.g., \cite{giltner_etal_1983} and Chapter 11 in \cite{storer_1988}). The apparent advantage of the SP system is the creation of more sophisticated grammars yielding higher levels of compression, and the application of the idea to kinds of information---such as moving images---where an ordinary dictionary would not work.

Since ordinary TVs work well without this kind of mechanism, some people may argue that there is no need for anything different. But:

\begin{itemize}

\item The growing popularity of video, TV and films on mobile services is putting pressure on mobile bandwidth \cite{note_15},
and increases in resolution will mean increasing demands for bandwidth via all transmission routes.

\item It is likely that some of the bandwidth for terrestrial TV will be transferred to mobile services \cite{note_16},
creating an incentive to use the remaining bandwidth efficiently.

\item There would be benefits in, for example, the transmission of information from a robot on Mars, or any other situation where a relatively large amount of information needs to be transmitted as quickly as possible over a relatively low-bandwidth channel.

\end{itemize}

\subsubsection{Potential for Gains in Efficiency in Processing and in the Use of Energy}

If, for example, we wish to search a large body of data, $I$, for instances of a pattern like ``Treaty on the Functioning of the European Union'', and to conduct similar searches for other patterns, there can be savings in processing if:

\begin{enumerate}

\item $I$ is compressed by finding recurrent patterns in $I$, like ``Treaty on the Functioning of the European Union'', and replacing each instance of a given pattern by a relatively short code such as ``TFEU'', or, more generally, compressing $I$ via grammar discovery and encoding as described in \cite{wolff_2006} (Chapter 9 and Section 3.5) and \cite{sp_extended_overview} (Sections 4 and 5).

\item Searching is done with ``TFEU'' instead of ``Treaty on the Functioning of the European Union'', or, more generally, searching is done with a compressed representation of the search pattern via the encoding processes in the SP system.

\end{enumerate}

Although the first ``encoding'' step requires processing, that one-time cost can be more than offset by consequent savings in many searches. The savings arise because of the reduced size of $I$ and because of the relatively small size of a code like ``TFEU'' compared with the pattern that it represents.

Since searching lies at the heart of how the SP system works, gains in the efficiency of searching should mean gains in the overall efficiency of processing in the SP system. There should be corresponding reductions in the use of energy.

\subsection{Medical Diagnosis}\label{medical_diagnosis_section}

The way in which the SP system may be applied in medical diagnosis is described in \cite{wolff_medical_diagnosis}. The expected benefits of the SP system in that area of application include:

\begin{itemize}

\item A format for representing diseases that is simple and intuitive.

\item An ability to cope with errors and uncertainties in diagnostic information.

\item The simplicity of storing statistical information as frequencies of occurrence of diseases.

\item The system provides a method for evaluating alternative diagnostic hypotheses that yields true~probabilities.

\item It is a framework that should facilitate the unsupervised learning of medical knowledge and the integration of medical diagnosis with other AI applications.

\end{itemize}

The main emphasis in \cite{wolff_medical_diagnosis} is on medical diagnosis as pattern recognition. But the SP system may also be applied to causal diagnosis (\cite{wolff_2006} (Section 7.9) and \cite{sp_extended_overview} (Section 10.5)) which, in a medical context, may enable the system to reason that ``The patient's fatigue may be caused by anemia which may be caused by a shortage of iron in the diet''.

\subsection{Managing ``Big Data'' and Gaining Value from It}\label{big_data_section}

The SP system has potential in the management and analysis of the large volumes of data that are now produced in many fields (``big data''), as discussed in \cite{sp_big_data} and summarized here.

The system may help overcoming the problem of variety in big data: the many different formats for knowledge and consequent problems in analysis. In view of the versatility of SP patterns, within the SP framework, to represent diverse kinds of knowledge \cite{sp_extended_overview} (Section 7), there is potential for them to serve as a universal format for knowledge and to facilitate the processing of diverse kinds of knowledge. The SP framework may also provide a mechanism for translating the original data into the universal format.

In the interpretation of data, the SP system has capabilities that include such things as the parsing and production of language, pattern recognition, and various kinds of probabilistic reasoning. But potentially the most useful facility with big data would be scanning for patterns, with recognition of family-resemblance or polythetic categories, at multiple levels of abstraction and with part-whole hierarchies, with inductive prediction and the calculation of associated probabilities, with a role for context in recognition, and robust in the face of errors of omission, commission or substitution. There are potential applications in several areas including, security, finance, meteorology, and astronomy.

With solutions to residual problems in unsupervised learning \cite{sp_extended_overview} (Sections 3.3 and 5.1.4), it seems likely that the SP machine, via the ``DONSVIC'' principle \cite{sp_extended_overview} (Section 5.2), will prove useful in discovering ``interesting'' or ``useful'' structures in big data, including significant entities, class-inclusion structures, part-whole structures, rules, regularities and associations, and discontinuous dependencies in data. The use of one simple format for different kinds of knowledge is likely to facilitate the learning of concepts that combine diverse kinds of knowledge.

Because of its potential for information compression (Section \ref{information_compression_section}), the SP system may achieve useful reductions in the volume of big data and thus facilitate its storage and management. There is potential for substantial economies in the transmission of big data (Section 6.7.1),
and for the efficient processing of compressed data, without the need for decompression (Section 6.7.2). And unlike conventional systems for compression of information, the SP system has capabilities in several aspects of intelligence.

\subsection{Other Areas of Application}

Some other areas of application where there is potential for the SP system are grouped here because, so far, that potential has not yet been much explored. But any of them may turn out to be significant.

\subsubsection{Knowledge, Reasoning, and the Semantic Web}\label{semantic_web_section}

The SP framework may contribute to the development of the ``semantic web''---a ``web of data'' to provide machine-understandable meaning for web pages \cite{bernerslee_etal_2001,shadbolt_hall_bernerslee_2006,note_17}.
In this connection, the main attractions of the system appear to be:

\begin{itemize}

\item {\em Simplicity, versatility and integration in the representation of knowledge}. The SP system combines simplicity in the underlying format for knowledge with the versatility to represent several different kinds of knowledge---and it facilitates the seamless integration of those different kinds of knowledge \cite{sp_extended_overview} (Section 7).

\item {\em Versatility in reasoning}. The system provides for several kinds of reasoning and their inter-working (Chapter 7 in \cite{wolff_2006} and Section 10 in \cite{sp_extended_overview}).

\item {\em Natural language understanding}. As suggested in Section \ref{nl_applications_section}, the SP system has potential for the understanding of natural language. If that proves possible, semantic structures may be derived from textual information in web pages, without the need, in those web pages, for the separate provision of ontologies or the like.

\item {\em Automatic learning of ontologies}. As suggested in \cite{sp_extended_overview} (Section 5.2), the SP system has potential, via the DONSVIC principle, for the extraction of interesting or useful structures from data. These may include the kinds of ontologies which have been a focus of interest in the development of the semantic web.

\item {\em Uncertainty and vagueness}. The SP system is inherently probabilistic and it is robust in the face of incomplete information and errors of commission or substitution. These capabilities appear promising as a means of coping with uncertainty and vagueness in the semantic web \cite{lukasiewicz_straccia_2008}.

\end{itemize}

\subsubsection{Bioinformatics}\label{bioinformatics_section}

As described in \cite{sp_extended_overview} (Section 4), a central idea in the SP system is the concept of {\em multiple alignment}, borrowed from bioinformatics but with important differences from how that concept is normally understood in research in biochemistry.

Notwithstanding those differences, or perhaps because of them, the SP system has potential for the analysis of DNA sequences and amino-acid sequences in any or all of the following ways:

\begin{itemize}

\item The formation of alignments amongst two or more sequences, much as in multiple alignment as currently understood in bioinformatics.

\item The discovery or recognition of recurrent patterns in DNA or amino-acid sequences, including discontinuous patterns.

\item The discovery of correlations between genes, or combinations of genes, and any or all of: diseases; signs or symptoms of diseases; or any other feature or combination of features of people, other animals, or plants.

\end{itemize}

\subsubsection{Maintaining Multiple Versions and Parts of a Document or Web Page}

The multiple alignment concept provides a means of maintaining multiple versions of any document or web page, such as, for example, versions in different languages. Parts which are shared amongst different versions---such as pictures or diagrams---may be kept separate from parts that vary amongst different versions---such as text in a specific language. As with other kinds of data, the framework provides the means of maintaining hierarchies of classes (with cross classification if that is required) and of maintaining part-whole hierarchies and their integration with class hierarchies \cite{sp_extended_overview} (Section 9.1).

Updates to high-level classes appear directly in all lower-level classes without the need for information to be copied amongst different versions.

Of course, these kinds of things can be done in other ways. The strength of the SP framework is its versatility, reducing the need for {\em ad hoc} solutions in different areas.

\subsubsection{Detection of Computer Viruses}\label{computer_viruses_section}

The detection of already-known computer viruses and other malware can be more subtle and difficult than simply looking for exact matches for the ``signatures'' of viruses. The offending code may be rather similar to perfectly legitimate code and it may be contained within a compressed (``packed'') executable~file.

Here, the SP system and its capabilities for pattern recognition and learning may prove useful:

\begin{itemize}

\item Recognition in the SP system is probabilistic, it does not depend on the presence or absence of any particular feature or combination of features, and it can cope with errors.

\item By compressing known viruses into a set of SP patterns, the system can reduce the amount of information needed to specify viruses and, as a consequence, it can reduce the amount of processing needed for the detection of viruses.

\end{itemize}

\subsubsection{Data Fusion}

There are many situations where it can be useful to merge or integrate two or more sources of information, normally when there is some commonality amongst the different sources. For example, in ecological research there may be a need to integrate different sources of information about the movement around the world of whales or other creatures; and in the management of a website, there may be a need to consolidate a set of web pages that are saying similar things.

At first sight, the SP system appears to be tailor made for this area of application. After all, the merging of fully or partially matching patterns lies at the heart of the SP system. But, as with big data (Section \ref{big_data_section}), the variety of different formats for knowledge can pose problems.

A possible solution, as with big data, is to translate different kinds of knowledge into a universal format. As before, the SP framework may provide that universal format and a mechanism for the~translation.

\subsubsection{New Kinds of Computer}\label{new_kinds_of_computer_section}

In research to develop new kinds of computer, such as optical or chemical computers, it is commonly assumed that the first step must be to create the optical or chemical equivalent of a transistor or logic gate, probably because of the significance of those kinds of structures in the evolution of electronic computers as we know them today.

But the SP theory, with its emphasis on information compression via the matching and unification of patterns, suggests new approaches to the development of new technologies for computing, perhaps by-passing some of the structures and concepts that have been prominent in the development of electronic~computers.

A potential benefit is that, with that primary focus on the matching and unification of patterns, it may prove easier to take advantage of the potential for high levels of parallelism in optical or chemical computers, or new kinds of electronic computers.

\subsubsection{Development of Scientific Theories}\label{science_section}

Although Occam's Razor is widely recognized as a touchstone of good science and, as noted earlier, John Barrow has written that ``Science is, at root, just the search for compression in the world.'' \cite{barrow_1992} (p. 247), there seems to have been relatively little interest---apart from some work on principles of ``minimum length encoding'' (see, for example, \cite{solomonoff_1964,wallace_boulton_1968,rissanen_1978})
---in making a connection between the practice of information compression and the development of scientific theories.

A possible reason is that most compression algorithms yield an encoding of the target data which is, from a human perspective, remarkably opaque: the output is normally smaller that the input---a useful result---but the encoding does not normally contain structures or relations that are meaningful for people. As noted in \cite{sp_extended_overview} (Section 13), this is probably because compression algorithms are normally designed to be quick-and-dirty, aiming for speed on low-powered computers, with a low priority for finding structures that yield high levels of compression; and, in accordance with the previously-mentioned DONSVIC principle, it appears to be those structures that are most meaningful for people.

Perhaps this does not matter. In a ``practicalities-not-insight'' approach to artificial intelligence that appears to be favored by some researchers, results that are useful in practice are given a higher priority than ``explanation'' or ``understanding'' \cite{note_19}.

With the SP system, there appears to be potential for something different and ultimately more acceptable as a way doing science: the possibility of creating a ``discovery machine'' \cite{kelly_hamm_2013} (p.~108{\em ff.}), deriving scientific theories from data, automatically or semi-automatically, and for obtaining analyses that people will regard as meaningful and insightful and which are also useful and practical. 
There is also potential for the quantitative evaluation of theories, as mentioned in Section \ref{sp_simplicity_power_section}.

In this area, reasons for optimism include:

\begin{itemize}

\item An ``up market'' emphasis in the SP program on finding structures that are good in terms of information compression.

\item The DONSVIC principle, mentioned above---the discovery of natural structures via information compression \cite{sp_extended_overview} (Section 5.2). Structures that are good in terms of information compression seem generally to be structures that people regard as natural and meaningful.

\item The versatility of the multiple alignment concept in the representation and processing of structures that are meaningful to people.

\end{itemize}

\section{Seamless Integration of Structures and Functions within and between Different Areas \\ of Application}\label{integration_section}

This section emphasizes again the importance of how the SP system may integrate structures and functions both within and between different areas of application. Instead of what can sometimes be awkward marriages between incompatible systems, there can be smooth interworking of, for example, pattern recognition, reasoning, and learning; of different kinds of reasoning; of different kinds of knowledge; and so on.

As previously noted, this facilitation of integration arises from: the use of one simple format for different kinds of knowledge; the use of one system---the multiple alignment framework---for the processing of knowledge; and the over-arching principle of information compression by the matching and unification of patterns.

As we have seen, the system provides for the integration of syntax and semantics in natural language processing; for the same kind of processing to achieve both the analysis and the production of language; for the integration of class-inclusion hierarchies, part-whole hierarchies, and other forms of knowledge; and for the integration of different aspects of intelligence in any combination: natural language processing, pattern recognition, different kinds of reasoning, unsupervised learning, and so on.

These kinds of integration are probably necessary to achieve human-like versatility and flexibility in the workings of computers.

\section{Conclusions}

The SP theory of intelligence combines conceptual simplicity with descriptive and explanatory power in several areas, including concepts of ``computing'', the representation of knowledge, natural language processing, pattern recognition, several kinds of reasoning, the storage and retrieval of information, planning and problem solving, unsupervised learning, information compression, neuroscience, and human perception and cognition.

In the SP machine there is potential for the simplification of computing systems, including software, with corresponding savings in the time, effort and cost in the development of applications, and other benefits related to information compression.

As a theory with a broad base of support, the SP theory promises deeper insights and better solutions in several areas of application, including unsupervised learning, natural language processing, autonomous robots, computer vision, intelligent databases, software engineering, information compression, medical diagnosis, and big data; and perhaps also in areas such as the semantic web, bioinformatics, structuring of documents, the detection of computer viruses, data fusion, new kinds of computer, and the development of scientific theories.

The SP theory should facilitate the integration of structures and functions, both within a given area and amongst different areas---a likely pre-requisite for the achievement of human-like versatility and flexibility in the way computers work.

These potential benefits and applications are not merely of theoretical interest. If, as a conservative estimate, they were to add 5\% to the value of annual worldwide IT investments \cite{note_21},
they would be worth \$190 billion each year, and increasing with the continuing growth of IT.

As suggested in \cite{sp_extended_overview} (Section 3.2) and in Section \ref{sp_summary_section} of this article, further development would be assisted by the creation of a high-parallel, open source version of the SP machine, available to researchers throughout the world to examine what can be done with the system and to create new versions of it.

\acknowledgements{Acknowledgements}

I am grateful to anonymous referees for many useful comments, and to Daniel Wolff for pointing out the potential of the SP system in the detection of computer viruses and in the management and analysis of big data.

\section*{\noindent Conflicts of Interest}

\vspace {12pt}
The author declares no conflict of interest.

\bibliographystyle{mdpi}
\makeatletter
\renewcommand\@biblabel[1]{#1. }
\makeatother


\begin{thebibliography}{----}
\bibitem[]{note_1}
According to this criterion, a good theory in a given area lies somewhere between things which are simple but very general---Such as the laws of arithmetic---And over-specific ``theories'' that simply repeat what it is they are meant to explain.
\bibitem[]{note_6}
Even in systems such as DBMSs or expert-system shells, mentioned above, the domain-specific knowledge which is loaded into the system may be regarded as part of the software.
\bibitem[]{note_8}
That number may be reduced if one or more languages are used as stepping stones to the translation of others.
\bibitem[]{note_9}
This reference to autonomous robots does not in any way endorse or defend the unethical or illegal use of airborne drones or other autonomous robots to cause death, injury, or damage to~property.
\bibitem[]{note_10}
As with information compression, a focus on the isolation of discrete objects in binocular vision is distinct from the more usual interest in the way that slight differences between the two images enables us to see the scene in depth.
\bibitem[]{note_14}
Notice that the distinction, just referenced, between domain-specific knowledge and MUP instructions is different from the often-made distinction between declarative programming and imperative or procedural programming. It is envisaged that the SP machine would reduce or eliminate the MUP instructions in software but that it would be able to model both the declarative ``what'' and (Section 6.6.1) the procedural ``how'' of the real world.
\end{thebibliography}



\end{document}